# Benchmarking Hebbian learning rules for associative memory

Anders Lansner[1,2], Naresh B Ravichandran[1], Andreas Knoblauch[3], Pawel Herman[1,4]

[1]KTH Royal Institute of Technology, EECS, Stockholm
[2]Stockholm University, [Dept. of??] Mathematics, Stockholm
[3]KEIM-Institute, Albstadt-Sigmaringen University, Germany
[4]Digital Futures, KTH Royal Institute of Technology, Stockholm

Corresponding author: Anders Lansner, ala@kth.se

**Abstract**

Associative memory or content addressable memory is an important component function in computer science and information processing, and at the same time it is a key concept in cognitive and computational brain science. Many different neural network architectures and learning rules have been proposed to model the brain's associative memory while investigating key component functions like pattern completion and rivalry together with noise reduction. A less investigated but equally important capability of active memory is prototype extraction where the training set comprises pattern instances generated by distorting prototype patterns and the task of the trained network is to recall the generating prototype given a new instance. In this paper we benchmark the associative memory function of seven different Hebbian learning rules employed in non-modular and modular recurrent networks with winner-take-all dynamic operating on moderately sparse binary patterns. Overall, we find that the modular networks have largest memory quantified as pattern storage capacity. The popular standard Hebb rule comes out with worst capacity while covariance learning proves to be robust but have low capacity, and the Bayesian-Hebbian rules show highest pattern storage capacity under the different conditions tested.

## Introduction

Associative memory as a concept in computer science refers to a memory that is content addressable, i.e., able to retrieve a stored item when given a fragment or distorted copy of it or to retrieve an item when cued by another associated item. Such so-called auto- and hetero-association, respectively, also reflects the meaning of associative memory in cognitive brain science and psychology. Associative memory capabilities of recurrent cortical neural networks are thought to underlie fundamental aspects of our brain's cognitive and perceptual functionality, e.g., figure-ground segmentation, long-term memory, perceptual completion and rivalry (Amit, 1990; Lansner, 2009). Key elements of stimulus-response behavior and associative chaining of thought processes may be described as heteroassociative, with a stimulus item associated to, e.g., an action or successor in a sequence. The search for the neural mechanisms underlying human associative memory dates back at least to Donald Hebb's cell assembly theory and hypotheses about mental representations in the form of cell assemblies and memory based on synaptic associative, i.e. "Hebbian", plasticity (Hebb, 1949).

The focus of this paper is on functional aspects of one-layer autoassociative memory networks and related neuroscientific theories and computational models. Following Palm (2013), the term "neural associative memory" (NAM) is adopted to distinguish such neurally oriented models of associative memory from non-network models often studied in cognitive science and from other common types of error-correction based artificial neural networks. Given the biological plausibility of Hebb's hypothesis about associative learning in the brain, we focus here on different variants of Hebbian learning rules proposed for associative memory. Our aim is to compare quantitatively by means of a set of benchmarks the associative memory performance of these different learning rules when employed in a single layer NAM network model. To emphasize and better reflect the influence of the learning rules employed, we aimed to use generic procedures with minimal network architectures and activation functions.

Recent extensions to the basic NAM we study here include mechanisms to generate hidden layers and higher order internal representations, for example predictive coding models (Tang et al., 2023), modern Hopfield networks (Krotov & Hopfield, 2021), and sparse quantal Hopfield networks (Alonso & Krichmar, 2024). These more complex models demonstrate significantly enhanced capabilities compared to the classical Hopfield network. However, since they rely heavily on additional mechanism beyond the single recurrently connected layer of neural units and Hebbian plasticity, they are considered beyond the scope of the investigations presented here.

## *Associative memory, pattern reconstruction and prototype extraction*

An autoassociative memory is content addressable in the sense that when stimulated with some input pattern the most similar, in some metric, among the stored patterns is recalled. In the linear "matrix memories" the patterns are simply vectors with real valued components. In the non-linear models, the patterns are binary and have components {0,1} or {-1,1} or with continuous-valued activation in the corresponding interval, e.g. [0,1]. Neuron spiking frequency when subject to sensory input may be considered as a confidence of the key stimulus being present (Meyniel et al., 2015). The pattern format that comes closest to this view is one with components in some interval between zero and maximal firing frequency, possibly normalized to [0,1]. It is thus somewhat surprising that much work in the NAM field has used and still uses a bipolar pattern activation function, possibly due to the strong influence from spin glass physics, as noted by Palm[1].

Proper function of the associative pattern processing in the neural network requires that the memory is not overloaded. When too many patterns are stored in a fixed size network, memory function typically breaks down and recalled patterns become distorted or may even be spurious, without obvious relation to any of the stored patterns. Central questions in the field have been and is still what learning rule and activation function gives the highest *pattern storage capacity and scaling* to large network sizes, as well as how to avoid the above mentioned "catastrophic forgetting" (Burgess et al., 1991). This is also the main subject of this work where we compare by computational experiments how this capacity depends on, in particular, the learning rule used.

---

[1]Günther Palm, 2013: "… probably due to the misleading symmetry assumption (symmetry with respect to sign change) that was imported from spin-glass physics. This prevented the use of binary {0, 1} activity values and the corresponding Hebb rule and the discovery of sparseness."

Another interesting but less studied operation of NAM:s is that of *prototype extraction*. It emerges in a basic form when training an associative memory network with a number of pattern instances generated from one of a set of prototype patterns by adding some form of distortion. When recall is tested with new instances, the memory is expected to recall the generating most similar prototype pattern, which itself was never presented to the network. Such an operation is closely related to clustering in data science and to concept and category formation in human cognition. Work on such learning in ANN has been scarce, but see e.g. Amari (1977), Lansner (1985, 1986) and recent modeling of such operations in ANN (McAlister et al., 2024; Ross et al., 2017; Tamosiunaite et al., 2022) and human concept formation (Fernandino et al., 2022).

## *Related work on Neural Associative Memory*

Hebb's work and publications in the 1940's inspired research in early theoretical and computational brain science as well as in engineering. One focus was on recurrent spiking neural network models and testing for emergence of Hebbian cell assemblies in biological tissue. As an example, early computer simulations by Rochester et al. (1956) failed to show that cell assemblies with sustained activity could form in a recurrently connected network of spiking model neurons[2]. Other early work was on associative memory in the hippocampus (Marr, 1971) and further development of models of memory function, associative memory, and concept formation followed (see e.g. Amari, 1977, 1989; Anderson et al., 1977; Nakano, 1972).

In the electronics and computer science domain the earliest associative memory work was by Steinbuch (Steinbuch & Piske, 1963). Steinbuch's LernMatrix was a binary or real valued crossbar associative network that took binary or normalized real valued vectors as input and produced a binary or real valued weight matrix during learning that was then used to generate output from new input. An important focus was hardware realization and several devices were produced and even used in applications. Kohonen developed further the Correlation matrix memory, quite related to the LernMatrix with real valued weights (Kohonen, 1972). Associative memory also originated early in the research community around holographic associative memories (Gabor, 1968; Longuet-Higgins, 1968). Work by Willshaw et al. (1969) developed further the concept of associative memory models with a binary weight matrix similar to the binary LernMatrix and it was followed by in depth analyses of the storage capacity and recall mechanisms of such NAM:s (Knoblauch, 2010, 2011, 2016; Knoblauch et al., 2010; Knoblauch & Palm, 2020; Knoblauch & Sommer, 2016; Palm, 1980, 2013; Schwenker et al., 1996).

The interest among theoretical physicist in brain modeling and associative memory in the form of attractor neural networks was spawned by the work of Little (1974) and later popularized by Hopfield (1982), who brought into focus the analogy between spin-glass physics and brain neurodynamics. This work has been further extended and elaborated by many researchers (see e.g. Amit et al., 1987; Kanter & Sompolinsky, 1986) with the occurrence of fixpoint, line- and chaotic attractors, and phase transitions in focus.

---

[2] This was achieved only later when the neuron properties were modelled after cortical pyamidal cells instead of spinal motor neurons (Lansner, 1986).

The Bayesian Confidence Propagation Neural Network (BCPNN) was first introduced in the late 1980's (Lansner & Ekeberg, 1987, 1989) and later developed with a modular architecture of hypercolumns and minicolumns (Johansson & Lansner, 2007; Lansner & Holst, 1996a; Sandberg et al., 2002). It has been used extensively to model cortical associative memory in non-spiking and spiking forms (Fiebig et al., 2020; Fiebig & Lansner, 2017; Lansner et al., 2013; Lundqvist et al., 2010a, 2011). BCPNN is related to the Potts neural network with "multi-state neurons", which was introduced in the late 1980's (Kanter, 1988) and later used as an associative memory (Mari & Treves, 1998; Naim et al., 2018). Interest in this kind of modular neural network architectures has recently risen in the context of quantum computing (see e.g. Fiorelli et al., 2022).

## Methods

### *Network architectures and learning rules*

Associative memory network models traditionally have a simple architecture often with just one recurrent layer. Here we used such a one-layer architecture with binary {0, 1} units operating on sparse distributed activity patterns. Our focus on sparse distributed patterns is partly motivated by data on the estimated activity levels of neurons in mammalian neocortex based on energy calculations and single unit recordings, indicating that typically less than 1 % of pyramidal cells are active at any instant (Lennie, 2003; Quiroga, 2012; Waydo et al., 2006).

Two types of network configurations were considered in our study: a non-modular one with N units, designated as "KofN", and a modular one having H modules ("hypercolumns") with M units ("minicolumns) each, designated as "HxM". Again, the latter configuration is compatible with evidence for such a modularization in primate neocortex (Kaas, 2013; Mountcastle, 1997; Opris & Casanova, 2014; Wallace et al., 2022). For this kind of modular networks, the partitioning of the $N$ network units could be done in many ways, but we here followed the "small world" scheme, i.e. $H = M = \sqrt{N}$, proposed for cortex by Braitenberg (1978), which produced moderatly sparse activity patterns.

The field update equation for the $j$:th neural units was:

$$h_j = b_j + \sum_{i}^{N} \pi_i w_{ij}$$

where $N$ is number of units, $h_j$ is the field, $b_j$ is the bias, $\pi_i$ is the presynaptic unit activity and $w_{ij}$ is the weight between presynaptic unit $i$ and postsynaptic unit $j$. For the non-modular network we used a k-winners-take-all (kWTA) activation function. In the modular network, each module used a local WTA (single winner) activation function. To be able to compare the KofN and HxM types of networks, K was chosen equal to H, which results in the same number of active units. Other more capable and biologically plausible schemes have been proposed and evaluated, but for our purpose here of comparing different learning rules we judged the simplest activation functions to be most appropriate. We also employed iterative updating, making the recurrent network operate as a so called attractor associative memory (Amit, 1990). Such a retrieval scheme also enhances storage capacity (Schwenker et al. 1996).

In our comparison we considered only local learning rules of a Hebbian correlation-based type, expressed by simple probabilistic measures of neuron activity and co-activity available at the synapse (Gerstner et al., 2014; Minai, 1997; Stuchlik, 2014). We also included learning rules that feature intrinsic plasticity, i.e. an activity dependent regulation of unit baseline activity, as has been described experimentally (Egorov et al., 2002). We compared seven different learning rules (see also Table 2):

1. The *Willshaw* learning rule proposed by Willshaw et al. (1969) based on earlier work by Steinbuch and later analyzed extensively by Palm et al., see e.g. Palm (2013);
2. The standard *Hebbian* learning rule, see e.g. (Amari, 1977);
3. The sparse *Hopfield* learning rule based on Hopfield (1982) for binary neural units, later adapted for sparse activity patterns ((Amari, 1989);
4. The *Covariance* learning rule proposed in 1988 by Tsodyks and Feigelman, adding a term to the Hopfield learning rule, making independently active units develop zero weights between them (Tsodyks & Feigelman, 1988);
5. The *Presynaptic Covariance* learning rule proposed in 1997 with the intent to improve storage capacity for correlated patterns (Minai, 1997);
6. The *Bayesian Confidence Propagation* (BCP) learning rule derived from Bayes rule, initially proposed by Lansner & Ekeberg (1989) and later adapted to a modular neural network architeture (Johansson & Lansner, 2007; Lansner & Holst, 1996);
7. The *Bayesian Optimal Memory* (BOM) learning rule derived from probabilistic Bayesian considerations assuming independent inputs (naïve Bayes) (Knoblauch, 2011).

Initially, we aimed to include also the Storkey learning rule which adds pre- and postsynaptic field terms to the Hopfield rule (Storkey, 1997). However, the inclusion of the presynaptic field to the synaptic weight update is local in a technical sense but violates the biologically motivated synapse locality constraint used here, since it uses the field of the presynaptic unit, which is not available at a biological synapse.

The Hebbian learning rules included here, with the exception of BOM, can be properly expressed in terms of activity and co-activity statistics (see e.g. Minai, 1997), and formulated as in Table 2. BOM does not quite fit in this scheme and is instead described in Appendix A.

| Counter equations | p-estimate equations |
|---|---|
| $c = \sum_k 1$ | - |
| $c_i = \sum_k x_i^{(k)}$ | $p_i = \max(c_i/c, \varepsilon)$ |
| $c_j = \sum_k x_j^{(k)}$ | $p_j = \max(c_i/c, \varepsilon)$ |
| $c_{ij} = \sum_k x_i^{(k)} x_j^{(k)}$ | $p_{ij} = \max(c_{ij}/c, \varepsilon^2)$ |

Table 1: Equations used to accumulate the amount of unit activations and co-activations during training, which gives the statistics ($p$-estimates) to calculate weights as in Table 2. $k$ indexes patterns. Here $x_i$ and $x_j$ are training pattern components and $\varepsilon$ was arbitrarily set to $10^{-7}$ for all learning rules except BCP where it was set to the lower bound of unit probabilities, $\varepsilon = 1/(c+1)$, following (Martinez Mayorquin, 2022).

In all cases, training was conducted incrementally (sample-by-sample) in one-shot mode, i.e. each input pattern was imposed once on unit activity during training, while $p$-estimates were calculated based on a simple batch frequentist approach, as detailed in Table 1.

| | **Abbrev.** | **bias** | **weight** | **Reference** |
|---|---|---|---|---|
| Willshaw | WILL | - | $1\ if\ p_{ij} > 0, 0$ otherwise | Willshaw et al. 1969 |
| Hebb | HEBB | - | $p_{ij}$ | Amari 1977 |
| Hopfield | HOPF | - | $p_{ij} - a[p_i + p_j] + a^2$ | Hopfield 1982, Amari 1989 |
| Covariance | COV | - | $p_{ij} - p_i p_j$ | Tsodyks and Feigelman 1988 |
| Presynaptic covariance | PRCOV | - | $\frac{p_{ij} - p_i p_j}{p_j}$ | Minai 1997 |
| Bayes Optimal Memory | BOM | See Appendix A | | Knoblauch 2011 |
| Bayesian Confidence Propagation | BCP | $\log p_j$ | $\log \frac{p_{ij}}{p_i p_j}$ | Lansner and Ekeberg 1989 |

Table 2: Equations for computing bias and weight values for different learning rules expressed in terms of activity and co-activity statistics calculated as in Table 1, and formulated as probabilistic $p$-estimates (cf. e.g. Minai, 1997). Here $a$ is fraction of ones in a pattern, i.e. activity density.

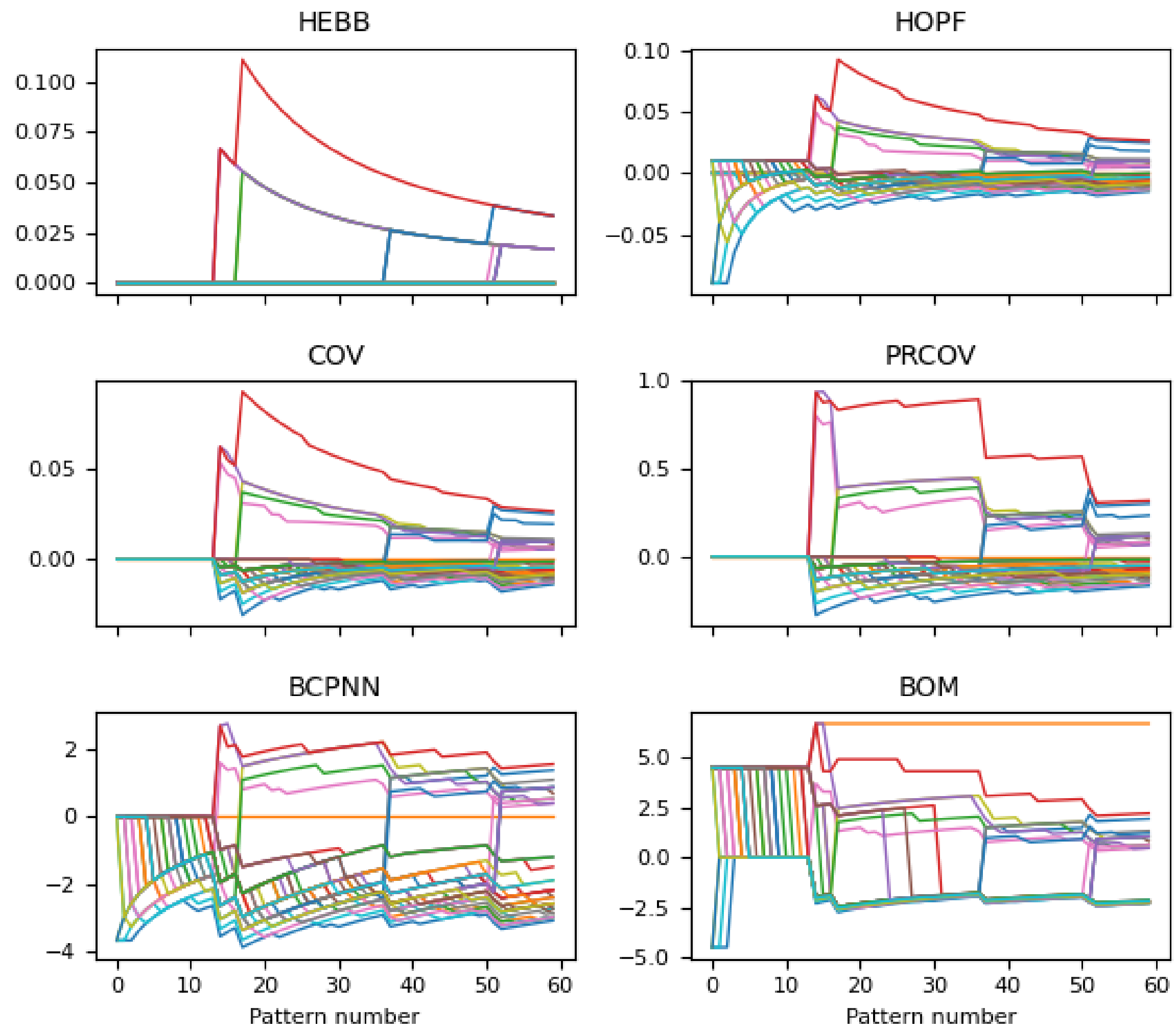

**Figure 1: Weight trajectories of different learning rules.** Different learning rules produce quite dissimilar weight trajectories when trained on the exact same training set of 60 patterns. Network format was 10x10 and the values of the same set of 40 weights are shown on the y-axis. The WILL learning rule with binary weights was excluded.

## *Experimental setup and evaluation*

The evaluation of the different learning rules was based on the network performance in two main types of benchmark problems: *pattern storage capacity* i.e. memory storage for binary patterns and *prototype extraction* from a set of distorted input pattern instances generated from binary prototype patterns. In both cases we investigated how the performance scaled with growing network size for non-modular and modular network configurations.

### Types of sparse random patterns

Three main types of sparse random patterns with subtypes depending on network architecture were used (Figure 2):

1. Standard binary patterns, designated “nrand” and “hrand”. Given the HxM format, one random unit was set active per hypercolumn and in the KofN format K active units were activated entirely randomly.

2. Silent binary patterns, designated “silent1”. A hypercolumn was designated as “silent” if it had the M’th unit set. Different fractions of silent hypercolumns (default 25%) were used. The same pattern format was used to test also the non-modular networks. Actually, for the BCP rule, earlier studies (unpubl.) have shown that it is possible to omit the marking of hypercolumns with an active “na” unit thus making the hypercolumns truly silent (input at 1/M for all M units) and the activity patterns truly sparse. However, this setup does not allow for use of the (k)WTA activation function and could thus not be further investigated here.
3. Correlated random patterns (Minai, 1997), with the correlation parameter $f_P$ (with default value of 0.1), designated “cnrand” and “chrand” respectively. Such correlations introduce both broadened distributions of unit usage and violations of the"naïve Bayes" assumption of independent pattern components.

The pattern distortion for creating training and test patterns as well as instances from pattern prototypes were introduced by resampling a specified fraction (default 10 %) of hypercolumns in a pattern. The constraints for silence and correlation were obeyed during such resampling. Since the number of hypercolumns is integer, for non-integer values of distortion (or silent fraction) the floor() and ceil() functions were mixed proportionally to achieve a proper mean.

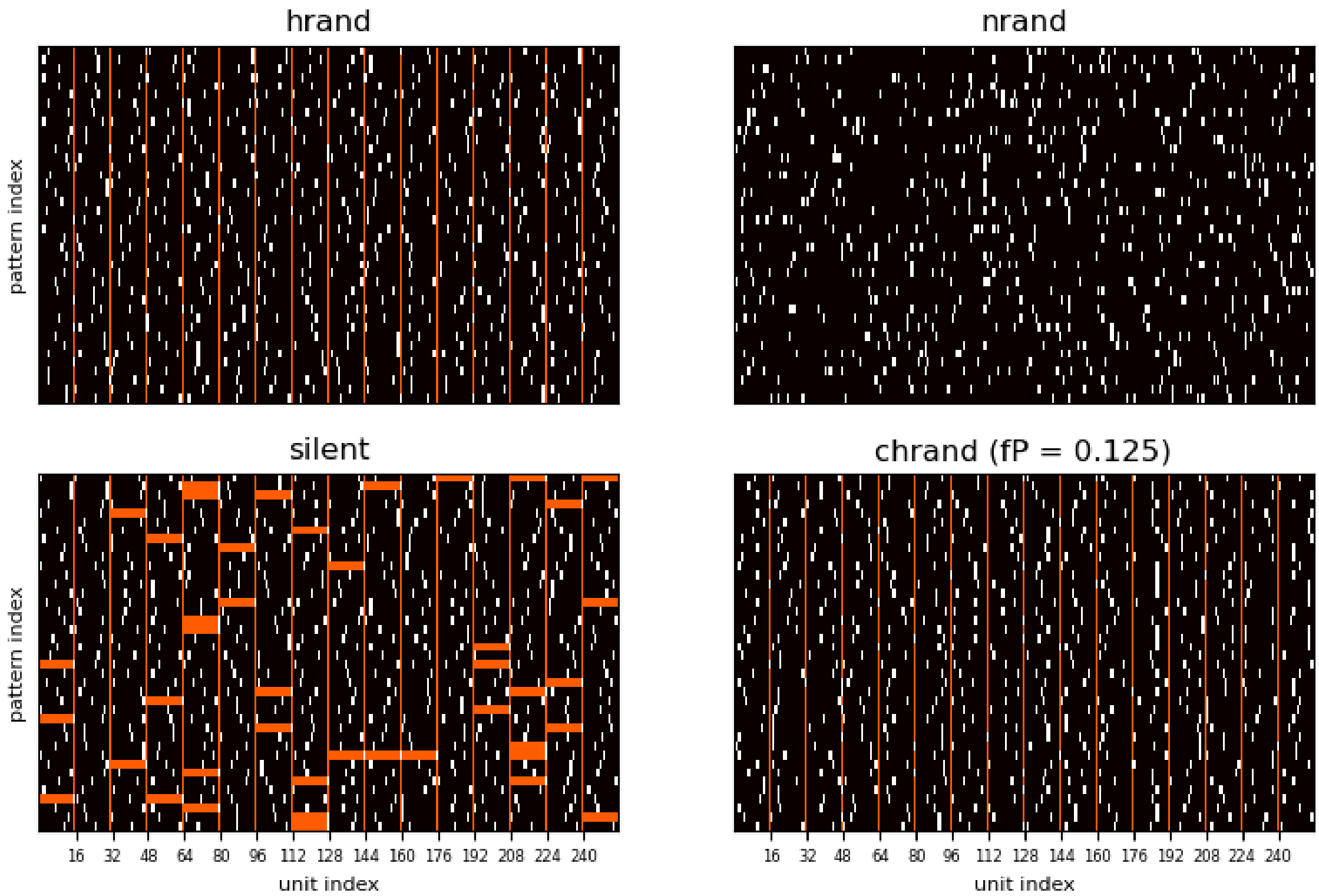

**Figure 2: Four different pattern types used.** Sets of 40 256-dimensional patterns are shown. The “silent hypercolumns” (marked orange) have their last unit set to 1. The “cnrand” type of patterns are not shown. For the modular patterns, vertical orange lines mark the last unit in hypercolumns.

The relevance of the silent pattern format (2 above) can be illustrated if we consider a database of different kinds of objects, characterized by a set of attributes, e.g. weight, length, colour, number-of-wheels, top-speed, number-of-legs, incubation time, etc. each discrete coded by one hypercolumn. Since for a specific object, only a fraction of all attributes would be relevant, the remaining ones would be “silent”, i.e. irrelevant. In a complex database, containing a wide variety of objects, the fraction of such irrelevant attributes could be quite large.It follows that this pattern format tests how sensitive the learning rules are to unbalanced unit activations, since the units marking silence has quite different statistics than the others.

Memory storage evaluation

Figure 3 exemplifies how the fraction of correct recall from distorted patterns behaves when the number of random patterns in the training set increases. For some time recall is perfect, but it eventually starts to fall down towards random performance. The result is qualitatively the same for a trained non-modular network. As seen from the error bars, the standard deviation is quite high among different networks storing the same number of patterns.

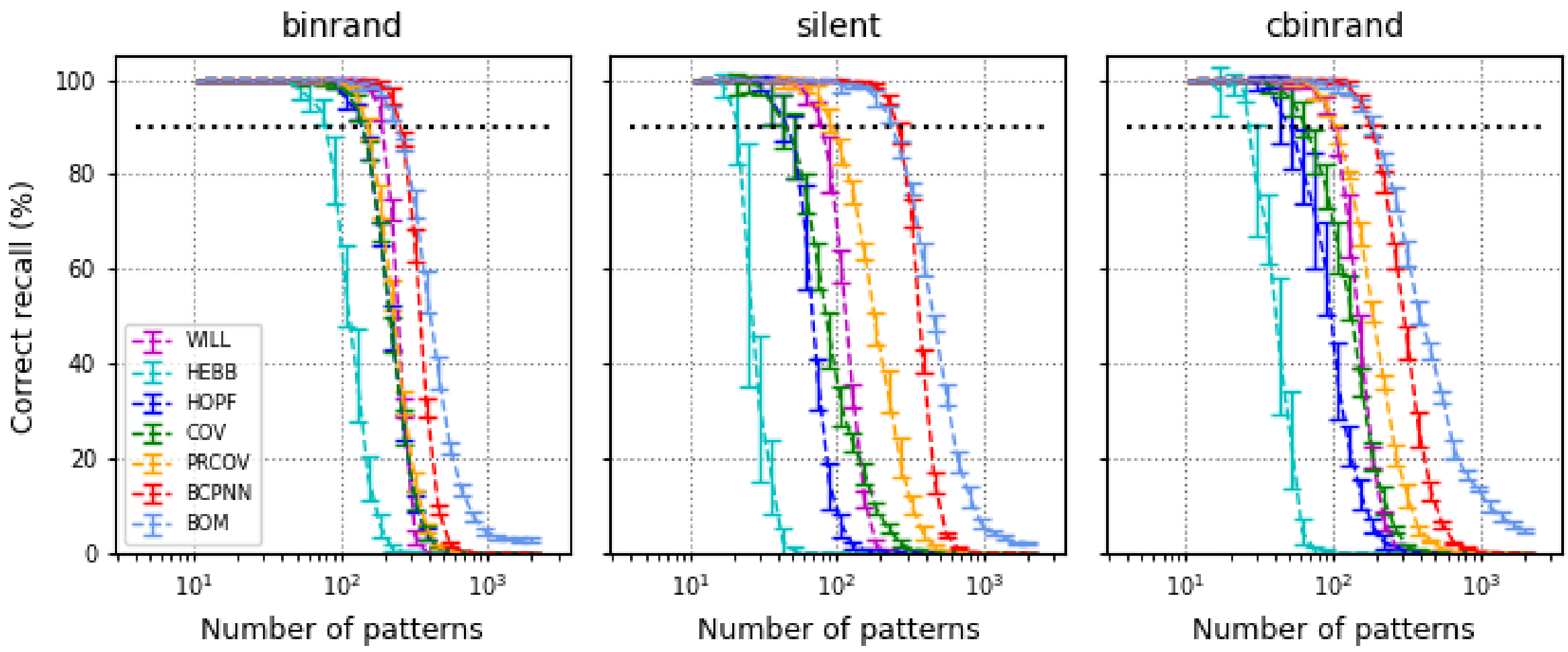

**Figure 3: Recall fraction of a modular network** (H = 16, M = 16) depending on pattern type and number of patterns in the training set for different learning rules. The network was trained with increasing number of patterns while measured recall fraction fell to low values. The dotted line marks 90% recall fraction. On average 10 % of hypercolumns were resampled in test patterns and error bars here show standard deviation. The number of test repeats per data point was 20.

The task of the network was to reconstruct the correct pattern from a test pattern with a specified fraction of hypercolumns distorted by resampling. Retrieval was iterative with the maximum number of iterations set to ten, but fewer were most often required to reach a stable attractor state. As stability criterion, it was checked in each step if the latest and next to latest activity states were the same. If so, iterations were terminated. Ten steps was reached in much less than 1% of runs and this small instability did not significantly affect reported recall performance.

Figure 3 demonstrates that different learning rules can store different number of patterns. The type of binary pattern also affects network performance significantly as can be seen in the second and third panel. To quantify the performance of the network in this pattern storage task, we defined capacity as the maximum number of patterns that could be stored while maintaining a perfect (correct) recall of 90% of the (distorted) test patterns, here designated as $P_{corr}$. This amounted to finding the number of patterns at which the curves in Figure 3 crossed the 90 % line. Since the process is probabilistic, a stochastic bisection method was used to estimate this crossing value (See Appendix B).

The number of binary patterns possible to store in a network is a commonly used measure of its pattern storage capacity. However, that number is highly dependent on network size and the activity density of the patterns, or more precisely, the information content of each pattern.

Patterns with few active units contain less information and can be stored reliably in higher numbers than more dense patterns. A complementary storage capacity measure in this context is the number of bits of information stored in the network relative to the number of free parameters (weights) or half of that for a symmetric weight matrix. This measure is illustrative to compare different learning rules and was estimated as follows.

Information based derivation of storage capacity scaling

Call the number of entries in the connection weight matrix of the network $N_W$ and the information stored per pattern $I_p$. Given an efficient memory structure, the number of patterns $P$ possible to store could be expected to scale as $N_W / I_p$ or for a symmetric W as $N_W/I_p/2$.

The total information contained in the recurrent weights is calculated as

$$I_W = \frac{N^2}{2} I_w \tag{1}$$

where $I_w$ represents the number of bits stored per weight. The number of bits per pattern stored is calculated as

$$I_p = \log_2 \binom{N}{K} \tag{2}$$

for the non-modular network type and as

$$I_p = H \log_2 M \tag{3}$$

for the modular networks. For such networks with $S$ silent hypercolumns we instead have

$$I_p = (H - S) \log_2 M + \log_2 \binom{H}{H-S} \tag{4}$$

From these relations, the scaling of $P$ can be calculated as $I_W/I_p$:

$$P = \frac{N^2}{2 \log_2 \binom{N}{K}} I_w \tag{5}$$

for the non-modular networks and

$$P = \frac{N^2}{2H \log_2 M} I_w \tag{6}$$

for the modular networks. If the relations $H = M = K = \sqrt{N}$, used for the respective network configuration are imposed, $I_w$ remains the only free parameter.

By fitting to the pattern storage capacity curves from our simulations using Eq. 5 and 6 we got a scaling relation for this capacity from which we could estimate $I_w$. Curve fits were done with the curve-fit() script in the python scipy package.

In the Results section we show that this method allowed us to estimate $I_w$ with good fit over the simulated range of $N$ and thus estimate how storage capacity scales with large $N$. Notably, under these conditions the estimate of $I_w$ was independent of $N$ but depended somewhat on the distortion level (see Fig. 7A) and the $P_{corr}$ threshold used.

Notably, the above analysis assumes that all patterns are retrieved without errors. This assumption holds only approximately, as we evaluate the networks at 90% correct level. So the acutal information per synapse will likely be a few percent smaller than estimated.

Prototype extraction and recall

For prototype extraction, several random prototype patterns were created and from each of them a number of distorted instances was then generated. Here, distorted patterns were used both for training and testing. The task of the network was to recall the generating prototype pattern from a previously unseen distorted version of it (Figure 4). As can be seen from the bottom row, the network was able to reconstruct one of the prototypes almost perfectly when trained with ten or more instances. The same prototype extraction happend for the nine other prototypes but is not shown. Notably, the calculation of the mean in the middle row was done given information of from which prototype a pattern instance was generated. This information was not given to the network, which therefore solved a harder problem.

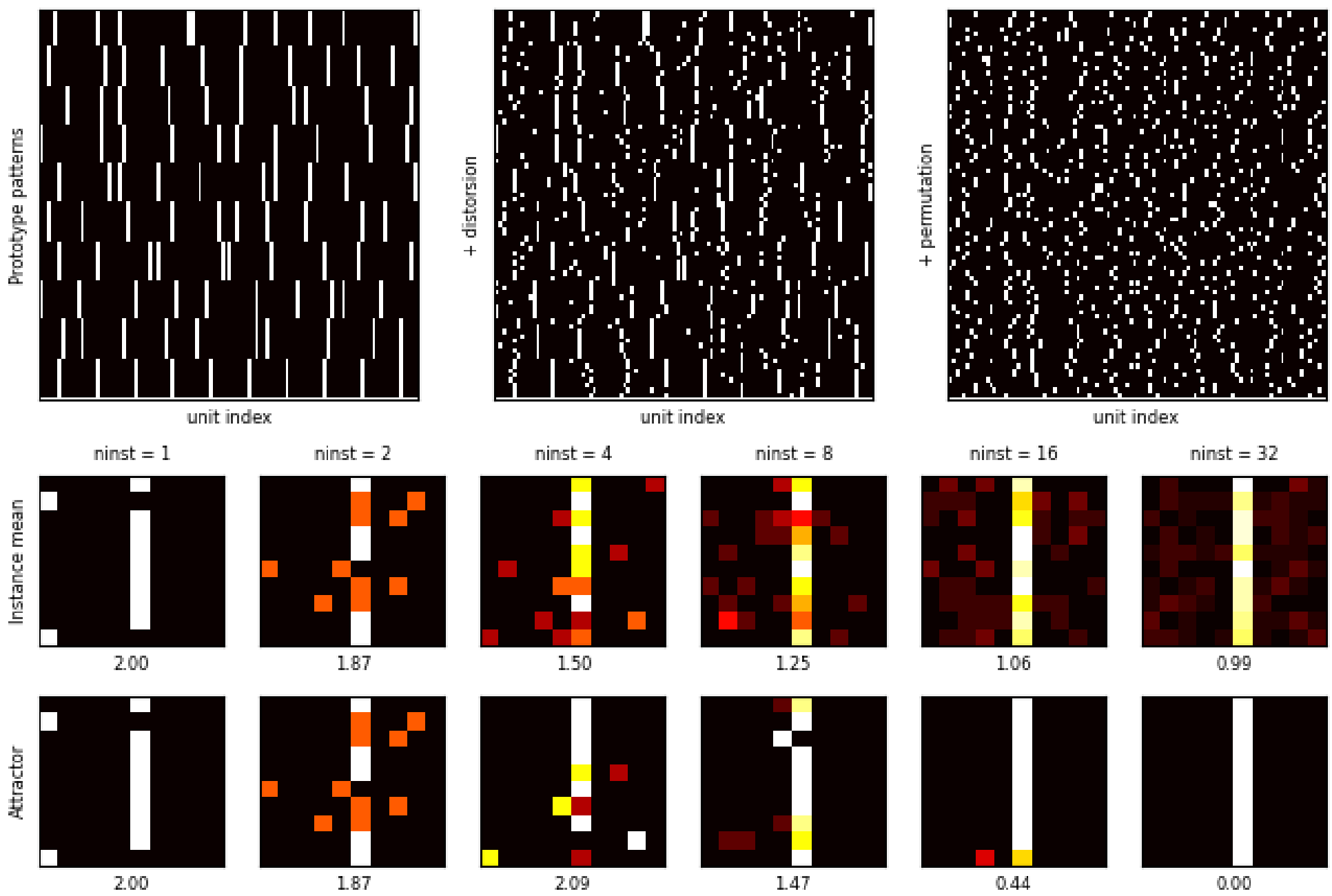

**Figure 4: Prototype extraction from different number of instances.** Top, left: 10 random 10x10 prototype patterns, one per row. Please note that the pattern on the last row, if unfolded in 2D forms a vertical bar. Top, middle: 32 training instances from each of the 10 prototypes. Each instance was resampled in 3 randomly selected hypercolumns. Top, right: The training instances in randomly permuted order forming the final training set. The panels in the second row show means of the 'ninst' number of training instances for the last prototype pattern. The bottom row shows the stable attractor state reached by a BCPNN trained with the 'ninst' instance patterns and tested with new such patterns. The number below each panel in the two lower rows gives the average Euclidean distance between the test and prototype pattern for the last prototype.

## Results

In this section we give results on pattern capacity and prototype extraction for the seven different learning rules and for different network architectures and pattern types. Results are mainly derived in the form of how the number of correctly recalled patterns/prototypes scale with network size. We further check the fit between the theoretical scaling estimate given above and our computational results as well as the impact of level of distortion and fraction of silent hypercolumns on pattern capacity given the different learning rules.

### *Storage capacity scaling*

Figure 5 shows how the pattern capacity at the 90% level scales with network size for non-modular (upper panels) and modular (lower panels) networks. It is evident that the storage capacity for different random pattern types varies with learning rule used and that some of these are very sensitive to the silent and correlated pattern types. Notably, the HEBB, HOPF and also the WILL learning rule with binary weights suffered quite dramatically from the silent and correlated pattern formats. The COV and PRCOV learning rules were quite robust but at a modest storage capacity.

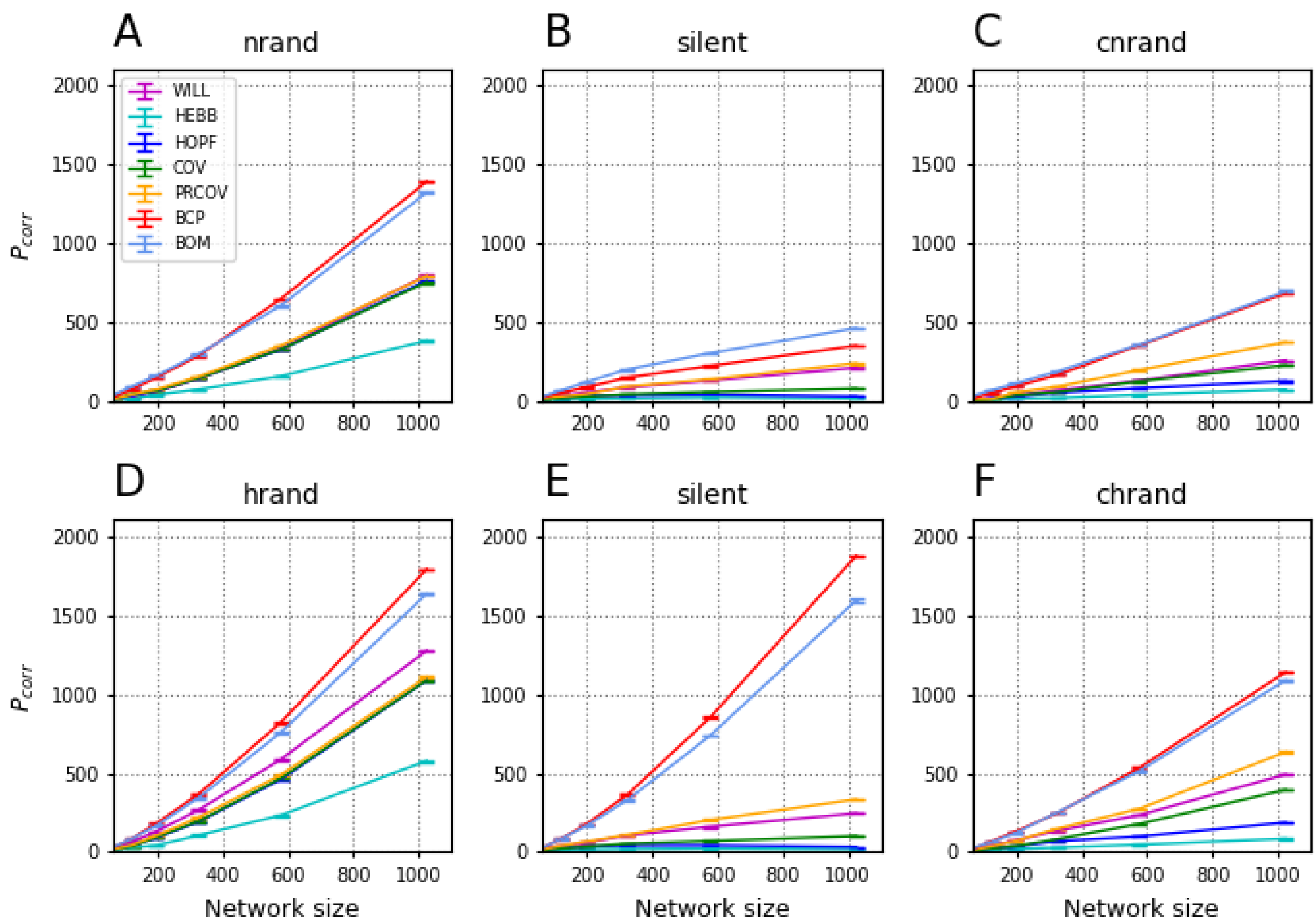

**Figure 5. Pattern capacity of non-modular and modular networks** depending on pattern type and learning rule, measured as the maximum number of patterns that allows for exact recall fraction above 90% ($P_{corr}$). The upper row corresponds to non-modular networks and the lower to modular networks. Each data point is the mean and standard deviation of five runs with different random seed. In the test patterns, 10 % of the hypercolumns were resampled. For the silent pattern format the fraction of silent hypercolumns was 25% and for the correlated pattern type the correlation parameter *fP* was 0.1. The legend in the upper left panel holds for all the panels. For table with data see Appendix C.

The BOM and BCP learning rules were best performing in all conditions. For non-modular networks they showed clearly decreased capacity for silent as well as correlated patterns. However, for modular networks they benefitted from the silent pattern format showing a higher pattern capacity than for standard random patterns ("hrand"). Indeed, silent patterns contain less information than the standard ones and should allow for a higher pattern capacity to maintain network information storage per weight constant. It is noteworthy that the Bayesian-Hebbian rules were at the top even for correlated patterns, with the PRCOV rule second, despite that the latter was developed to optimize performance on correlated patterns.

A surprising finding was that the BCP rule in several cases showed somewhat higher pattern capacity than the BOM rule, which was rigorously formulated to be Bayes optimal given independent input components and one-step retrieval. This anomaly could, however, be explained by the fact that optimality of BOM was proven for one-step retrieval when the input noise is properly set, but in later steps of iterative retrieval when the input noise is lower the parameter settings are no longer correct (Knoblauch, 2024). Then BCP can catch up and surpass BOM. But the capacity difference is still small compared to the gap to the others learning rules.

### *Fitting storage capacity scaling relations*

The theoretical relations given in Eqs. 5 and 6 in the Methods section were used to estimate $I_w$ from the data in Figure 5 for the standard random binary pattern types (hrand, nrand). As shown in Figure 6, the fit was quite good. From this fit we got the $I_w$ values for each learning rule and network type (Table 3). As can be seen, the numbers are rather similar for the two network types, though the actual number of patterns possible to store are significantly different due to the difference in information content of patterns in the two network architectures.

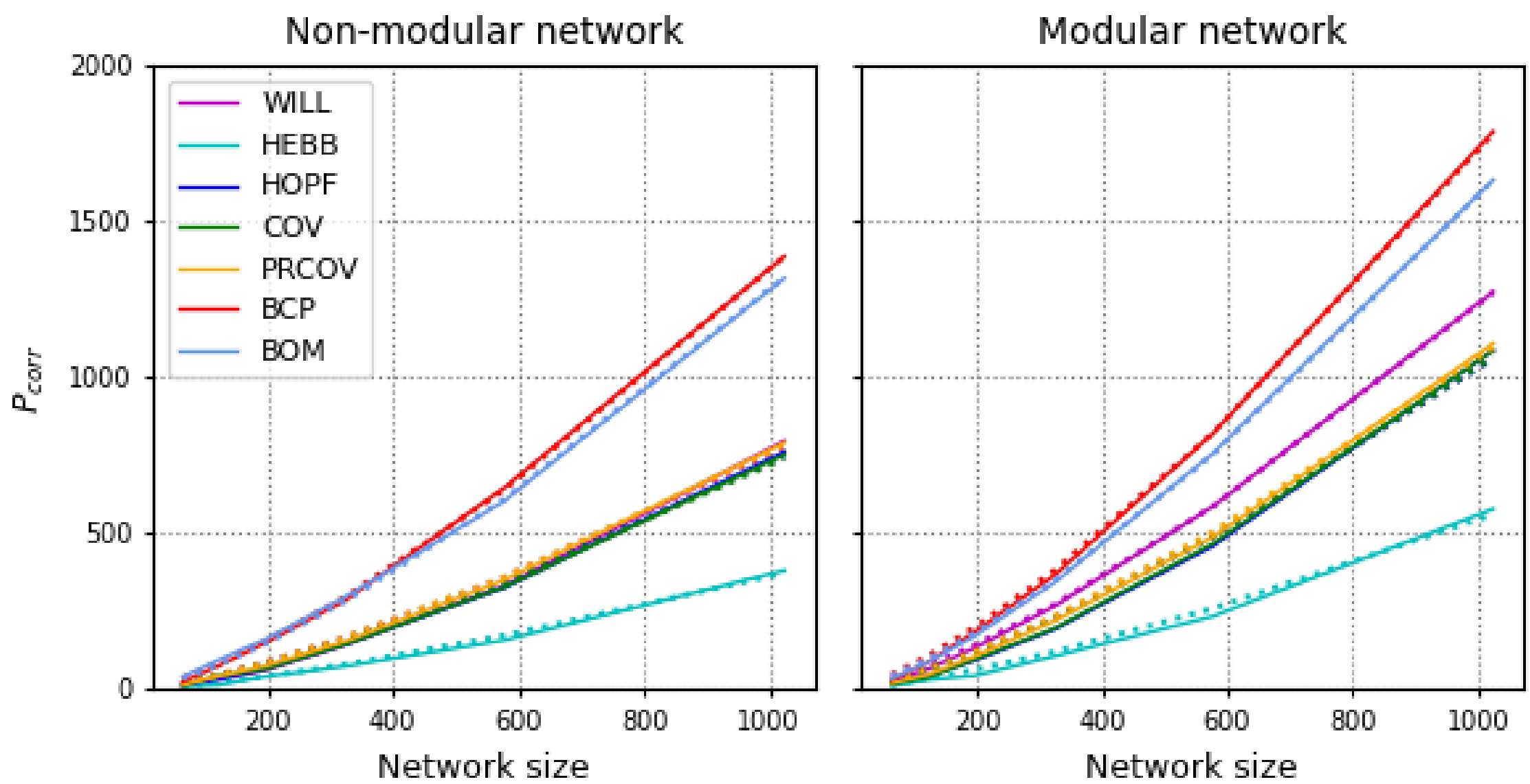

**Figure 6: Fitted pattern storage capacity scaling curves.** Pattern storage capacity scaling curves were fitted (dotted lines) using Eqs. 5 (left panel, non-modular network) and 6 (right panel, modular network) in the Methods section. This served as the basis for estimating the bits per weight, $I_w$ (values shown in Table 3).

| | WILL | HEBB | HOPF | COV | PRCOV | BCP | BOM |
|---|---|---|---|---|---|---|---|
| Non-modular | 0.24 | 0.12 | 0.23 | 0.23 | 0.24 | 0.42 | 0.40 |
| Modular | 0.39 | 0.17 | 0.32 | 0.33 | 0.33 | 0.54 | 0.50 |

Table 3: Estimated information in bits stored per weight. Data refer to the hrand/nrand pattern types. Note that BOM leads to an asymmetric weight matrix, which may decrease stored information per weight by factor 0.5, depending on the implementation (Appendix A).

### *Effect of fraction of distorted and silent hypercolumns*

We explored further the pattern storage capacity scaling with the different learning rules while changing the amount of test sample distortion and the fraction of silent hypercolumns in the patterns stored (Figure 7). The dependency of measured capacity with different levels of distorted (resampled) hypercolumns was investigated only for BCP, as it demonstrated the highest pattern storage capacity. As shown in Figure 7A, the capacity increased gradually with less distortion ending at 1892 stored patterns corresponding to 0.58 bits per weight.

Regarding the dependency of this capacity on fraction of silent hypercolumns (Figure 7B), the non-Bayesian rules all failed to store more patterns with higher numbers. The BCP rule performed well over the entire range. Notably, it maintained the same bits per weight value over a large range. The BOM rule showed an intermediate sensitivity and did not maintain the bits per weight value.

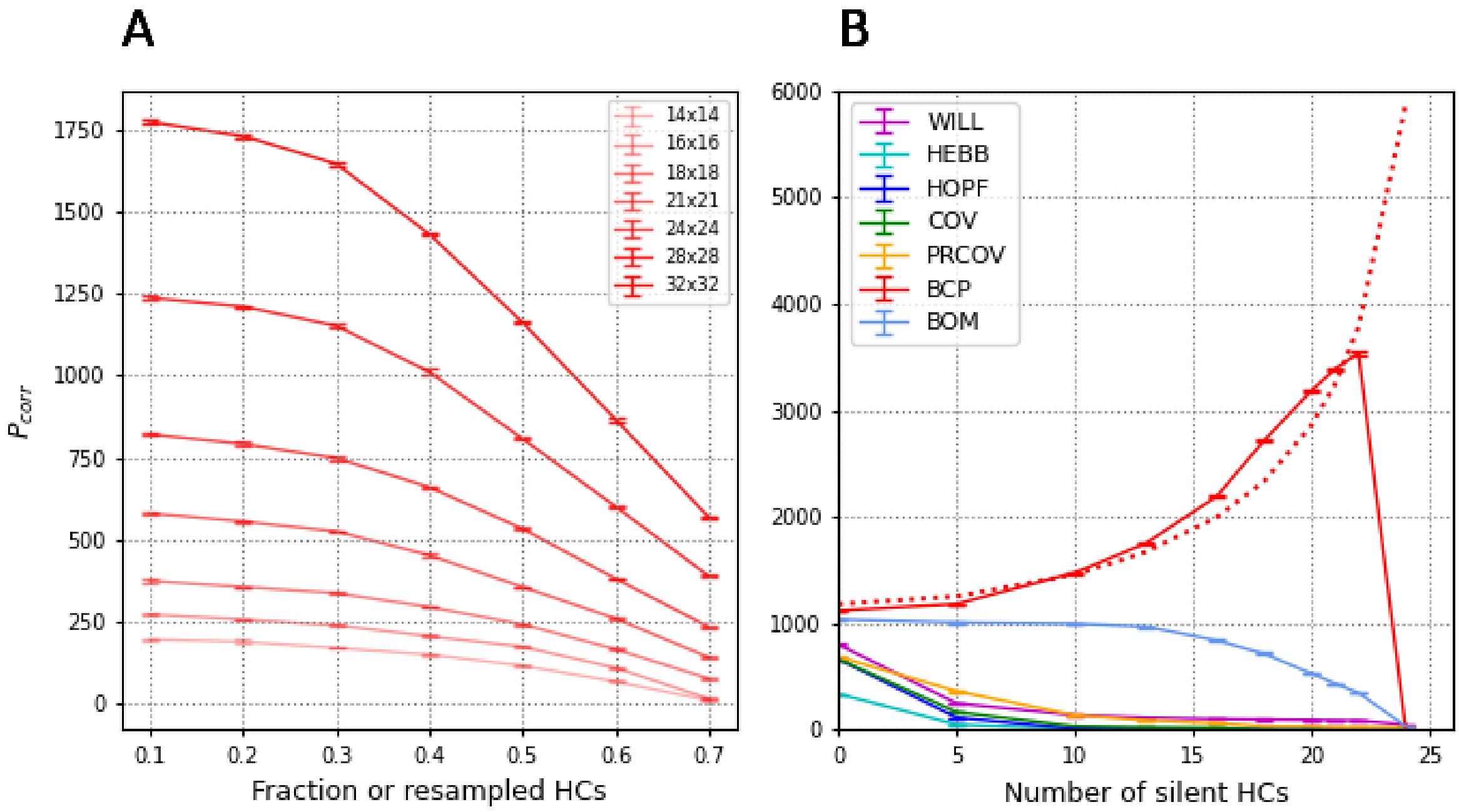

**Figure 7 A: Storage capacity at different amount of pattern distortion** shown for different network sizes using the BCP learning rule. Standard deviation was calculated from three runs with different random seed. **B: Recall at high fractions of silent hypercolumns.** A 19x19 network was trained with patterns with increasing fraction of silent hypercolumns using the different learning rules. 10% of the non-silent hypercolumns were distorted in the test patterns. Average and standard deviation of three runs is shown. The dotted line shows the capacity predicted from maintaining a constant bits per weight value using Eq. 4 in the Methods section.

### *Prototype extraction*

The prototype extraction capabilities of different network architectures and learning rules were evaluated in a similar manner as for storage of individual patterns. The main difference was that instead of training with just a number of patterns, it was done with instances generated by distortion from a set of training patterns (the prototypes) as described in the Methods section. The networks were tested for ability to reconstruct the generating prototype from unseen distorted test patterns.

Figure 8 illustrates how the pattern storage capacity for prototype patterns depended on the number of prototypes and instances per prototype for a fixed-size network (H = 16, M = 16) using the BCP learning rule. As can be seen, performace was steadily increasing with higher number of instances and only twenty instances were required to approach highest performance. The maximum number of prototypes possible to extract and store is, however, lower than in the reference scenario where the prototypes themselves are given during training (Figure 8, dashed curve). Another feature of this setup is the bimodal performance with the lowest score at an intermediate number of prototypes. We speculate that the observed nonmonotonicity originates from an interaction between the decreasing stability of individual memory patterns forming the prototypes and the corresponding increasing number of spurious stable states.

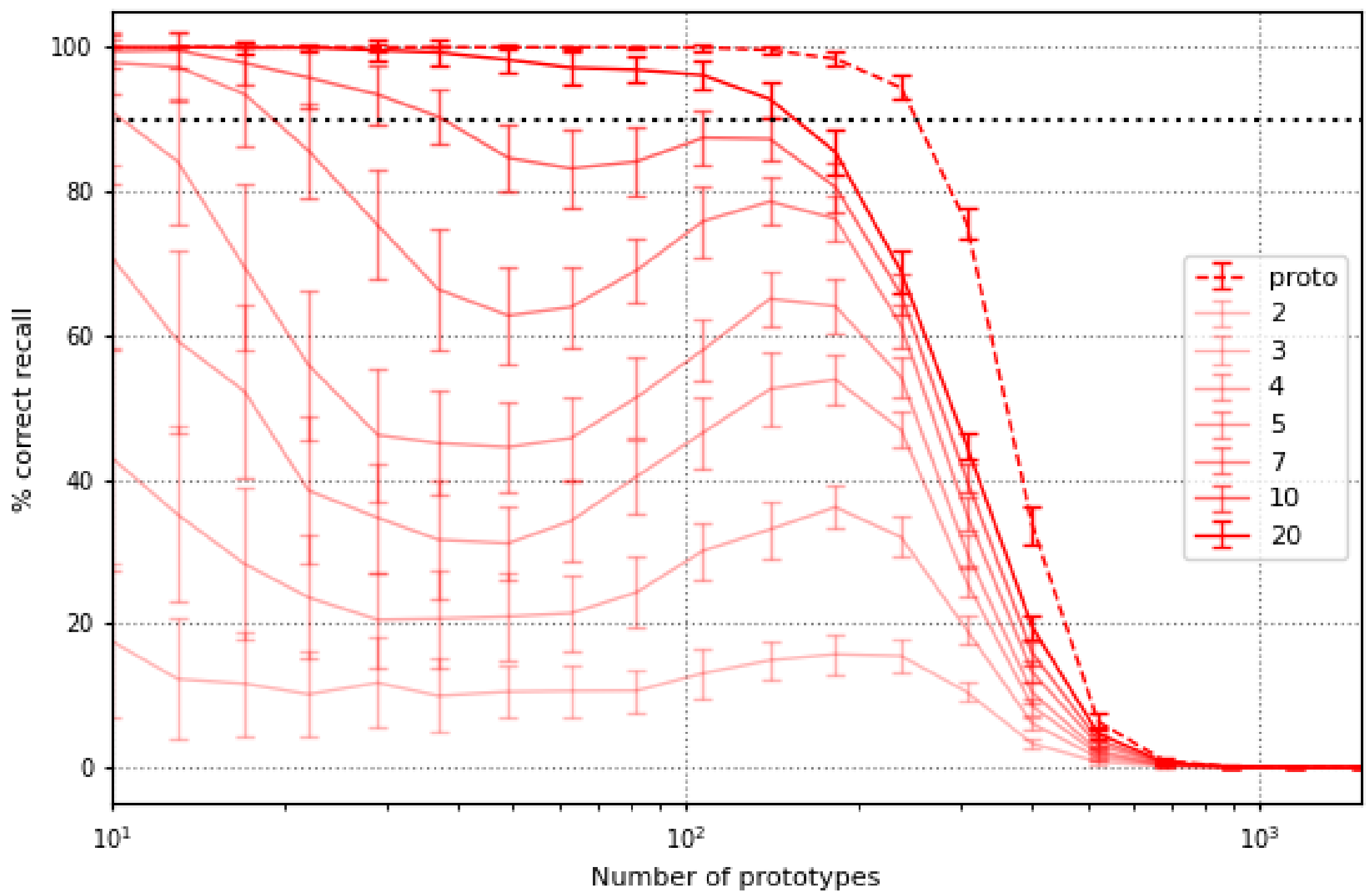

**Figure 8: Fraction of correct recall as function of number of prototypes and training instances.** Pattern capacity of a modular 16x16 BCPNN when trained with variable number of prototype patterns and number of instances generated from those. The task was to recall the generating prototype exactly. The dashed black line shows performance when only the prototype patterns themselves were stored. The dotted line marks the 90 % recall fraction and error bars show standard deviation from 20 repetitions.

Figure 9 shows how the prototype pattern storage capacity scaled with network size for the the two network architectures, the seven learning rules and different pattern types. As can be seen, the task was generally harder and performance overall lower than when storing individual patterns. It is not surprising that the WILL learning rule failed entirely due to the fact that a single pre/postsynaptic coincidence is sufficient to switch the synapse from 0 to 1 (the "zip-net" version of the WILL learning rule would remedy this problem (Knoblauch, 2010)). Though using real valued weights, as for the other learning rules, does not increase storage capacity *per se*, it provides a prototype extraction capability and robustness. Even for the real valued weight matrixes, bit precision could likely be lowered significantly, though this was not investigated further here (Vogginger et al., 2015).

For non-modular networks, the silent and correlated pattern types gave overall very low performance, with BOM and PRCOV showing some robustness. For modular networks, the picture was quite similar but capacity lower compared to the case with storing individual patterns. The Bayesian-Hebbian rules again showed highest storage capacity.

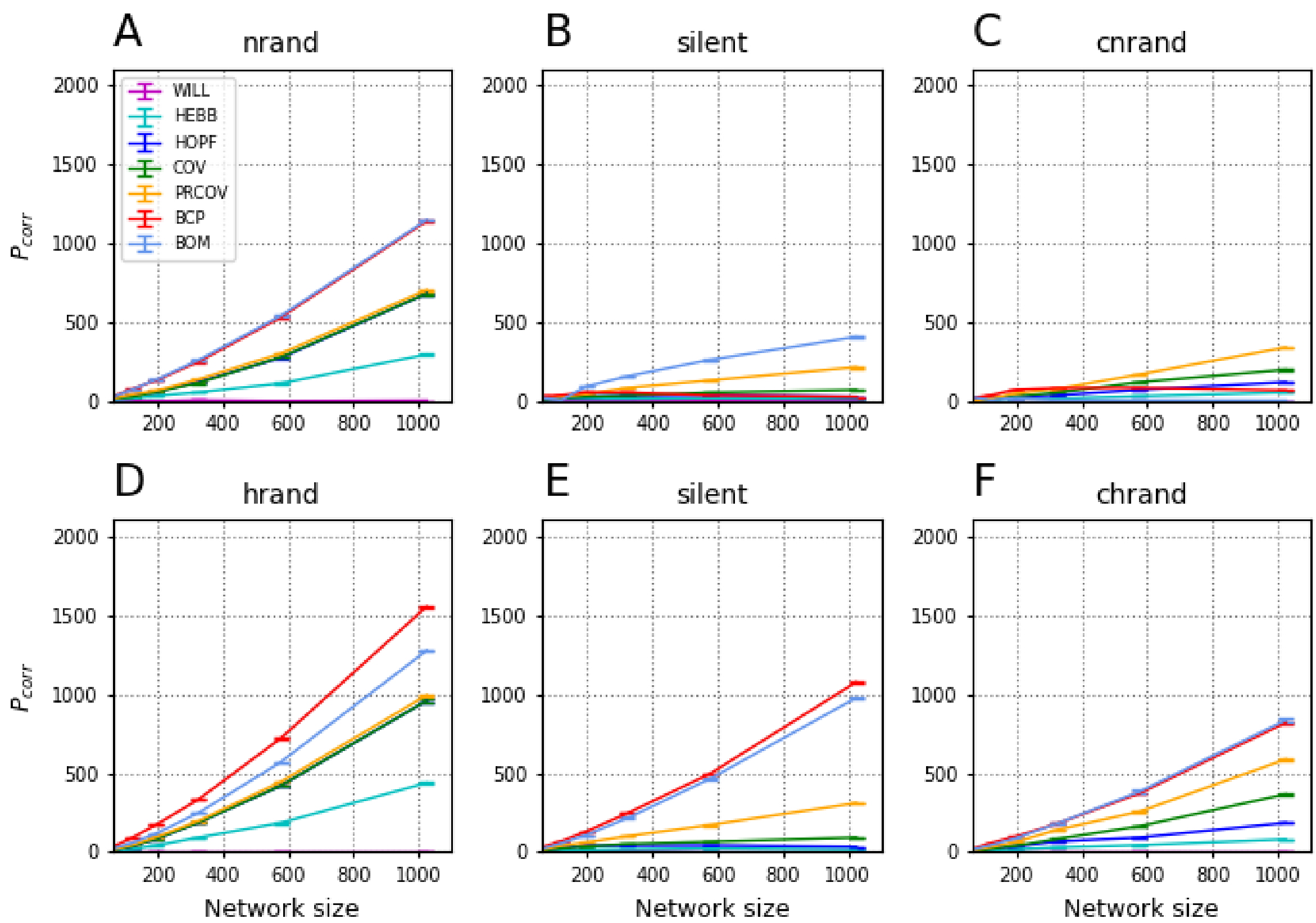

**Figure 9: Prototype storage capacity.** Performance measured as the maximal number of propotypes allowing exact prototype recall at 90% depending on type of prototype pattern and learning rule. Upper row refers to non-modular networks, lower row to modular networks. The number of training instances generated from each prototype was 20 and 10% of hypercolumns were resampled in both training and test instances. Legend in upper row left panel holds for all panels. For table with data see Appendix C.

## Discussion

In this work we have compared quantitatively by means of extensive computer simulations seven different learning rules with regard to associative memory pattern processing capabilities in terms of pattern storage capacity and prototype extraction. We used three different sparse random pattern types and two network architectures, non-modular and modular. Generally, the difference between non-modular and modular architectures was moderate. They stored a comparable amount of information when measured for the standard random patterns though somewhat fewer patterns in the non-modular case with higher information content per pattern.

Pattern storage capacity varied considerably with the learning rule used. The overall outcome was that the two Bayesian-Hebbian learning rules were superior with quite a large margin, to the others (Figures 5 and 9). In these benchmarks, the worst performing was the HEBB rule, which at the same time is the most popular in studies of associative memory. Our results differ from many previous theorectically derived and simulation based reported storage capacity measures, likely explained by the fact that our brain-compatible testing setup with moderately sparse random binary patterns differ from the often assumed dense or infinitely sparse patterns with bipolar unit activation.

For prototype extraction, the picture was somewhat different. Firstly, the task was generally harder than storage of individual patterns and for non-modular networks the silent and correlated pattern types gave overall very low performance. On the other hand, modular networks showed more robustness and as for individual pattern capacity, the Bayesian-Hebbian rules showed stable and significantly better performance than the others.

The results over all benchmark runs for the largest network size are summarized in Table 4. It can be seen that the mean pattern capacity (rightmost column) of the Bayesian-Hebbian learning rules (BCP and BOM) is close to 2x that of the runner up, which is PRCOV. Part of an explanation for the superiority of the Bayesian rules might be that they were derived from a naïve Bayes formalism for probabilistic inference and, in contrast to the others, operate in log space combining evidence multiplicatively rather than additively. For random patterns, the underlying independence assumption for naïve Bayes is likely to hold quite well, but this approach often works well even with real world data (Hand & Yu, 2001).

| | **Storage capacity scaling** | | | | | | **Prototype extraction** | | | | | | |
|---|---|---|---|---|---|---|---|---|---|---|---|---|---|
| | Non-modular | | | Modular | | | Non-modular | | | Modular | | | Learning rule mean |
| | nrand | silent | cnrand | hrand | silent | chrand | nrand | silent | cnrand | hrand | silent | chrand | |
| WILL | 796 | 210 | 258 | 1275 | 247 | 492 | 3 | 0 | 0 | 0 | 0 | 0 | 273 |
| HEBB | 381 | 17 | 74 | 578 | 18 | 86 | 294 | 14 | 54 | 438 | 16 | 82 | 171 |
| HOPF | 762 | 30 | 126 | 1088 | 34 | 188 | 680 | 26 | 119 | 958 | 34 | 186 | 353 |
| COV | 752 | 80 | 226 | 1088 | 102 | 395 | 680 | 70 | 198 | 958 | 93 | 367 | 417 |
| PRCOV | 791 | 239 | 372 | 1109 | 335 | 634 | 702 | 214 | 339 | 993 | 311 | 588 | 552 |
| BCP | 1388 | 344 | 682 | 1790 | 1875 | 1140 | 1137 | 26 | 67 | 1553 | 1076 | 820 | 992 |
| BOM | 1318 | 461 | 694 | 1634 | 1593 | 1088 | 1145 | 406 | 3 | 1275 | 976 | 836 | 952 |
| Mean | | | 476 | | | 802 | | | 275 | | | 550 | |

Table 4: Summary of learning rule performance over all conditions for $N = 1024$ and $H/K = 32$. The table shows the number of patterns stored given a 90% recall criterion, collected from the previously analysed simulation results. The average number per learning rule is given in the framed last column. The last row shows averages over all learning rules and pattern types for modular and non-modular architectures separately for pattern capacity and prototype extraction.

The last row of Table 4 shows further that the modular architecture overall gave higher pattern capacity in these benchmarks than the non-modular one. Furthermore, from a neurobiological point of view, the modular network seems straightforward to realize with local lateral inhibition and divisive normalization provided by basket cells (Carandini et al., 1997; Lundqvist et al., 2010b), whereas the kWTA selection of maximally active units over a network comprising many hypercolumns is more problematic to map to neocortical architecture.

Regarding parameter sensitivity there were not many learning rule related parameters in our benchmarking setup. Yet, one structural parameter fixed in our investigation was the number and size of hypercolumns in the modular network, at $\sqrt{N}$. This partitioning scheme was suggested by Braitenberg (1978) on the basis of cortical architecture, though with different types of modules in mind. As demonstrated here, this scheme works nicely in simulations of small to medium scale network models. However, many other HxM configurations may be relevant in different applications and dependence on this parameter remains to be investigated.

Moreover, our $\sqrt{N}$ scaling of HxM does not fit well with biological cortex sizes. The estimated number of minicolumns per hypercolumn in mammalian cortex is on the order of hundreds, so the number of hypercolumns would scale to much larger values, on the order of a million for a human sized neocortex. Uncertainty about the actual activity density in higher order cortex is a further complication. Possibly, the number of silent hypercolumns is quite high, thus resulting in very sparse and low information patterns (Lennie, 2003; Quiroga, 2012; Waydo et al., 2006) and a higher pattern storage capacity (Figure 7B). Another important but uncertain parameter in brain-scale networks is the density of connectivity between minicolumns, which should be one or two orders of magnitude higher than that between single neurons in the neocortex, but still far below the 100 % used in this study. Such dilution of connectivity would obviously reduce the storage capacities seen in this study.

## Conclusions

Our benchmarking study has demonstrated that a modular network architecture typically has higher pattern storage capacity than the non-modular network for individual patterns as well as for prototypes extraction given moderately sparse random binary patterns. Furthermore, the popular standard Hebbian learning rule came out with lowest capacity while the Bayesian-Hebbian learning rules came out on top with BCPNN performing on par with BOM, developed as a Bayes optimal memory.

**Acknowledgements**

Funding for the work was received from the Swedish e-Science Research Centre (SeRC), Digital Futures, Swedish Research Council (VR2018-05360 and VR2016-05871), the European Commission Directorate-General for Communication Networks, Content and Technology grant no. 101135809 (EXTRA-BRAIN) and Indo-Swedish Joint Network 2018 grant No. 2018–07079,.

Appendix A – The Bayes Optimal Memory learning rule (BOM)

The BOM learning rule minimizes output noise and maximizes storage capacity by activating neurons based on Bayesian maximum likelihood decisions [Knoblauch, 2011, 2024]. In the auto-associative case, the task is to store $M$ activity patterns $u^{\mu}$, where $\mu = 1, \dots, M$. Here the $u^{\mu} \in \{0,1\}^n$ are binary vectors of size $n$. Associations are stored in first order (neural) and second order (synaptic) counter variables

$$M_u(j) := \left|\{\mu : u_j^{\mu} = u\}\right|$$

$$M_{uv}(ij) := \left|\{\mu : u_i^{\mu} = u, u_j^{\mu} = v\}\right|$$

where $u, v \in \{0,1\}$ and $i, j \in \{1,2, \dots, n\}$. Note M=c, $M_1(i) = c_i$, and $M_{11}(ij) = c_{ij}$ for the counter variables of Table 1. Note that it is sufficient to store only $M$, $M_1$, and $M_{11}$, as all other counters can be reconstructed from $M_0(j) = M - M_1(j)$ and $M_{uv}(ij) = M_v(j) - M_{(1-u)v}(ij)$. Thus, exploiting the symmetry $M_{uv}(ij) = M_{vu}(ji)$, memory-efficient storage of the counter variables requires only $\left(1 + n + \frac{n(n+1)}{2}\right)\log_2 M \simeq \frac{n^2}{2}\log_2 M$ bits. For retrieval of stored memories we assume noisy query patterns $\tilde{u}$ resembling one of the stored memories $u^{\mu}$, where

$$p_{uv|w}(ij) := \mathrm{pr}\left[\tilde{u}_i = v | u_i^{\mu} = u, u_j^{\mu} = w\right]$$

defines "noise" as the probability of a bit switching from u to v. With this, biases $b_j$ and synaptic weights $w_{ij}$(from neuron $i$ to neuron $j$) of BOM write

$$b_j := (n-1)\log\frac{M_0}{M_1} + \sum_{i=1}^{n} \log\frac{M_{01}(1 - p_{01|1}) + M_{11}p_{10|1}}{M_{00}(1 - p_{01|0}) + M_{10}p_{10|0}}$$

$$w_{ij} := \log\frac{\left(M_{11}(1 - p_{10|1}) + M_{01}p_{01|1}\right)\cdot\left(M_{00}(1 - p_{01|0}) + M_{10}p_{10|0}\right)}{\left(M_{10}(1 - p_{10|0}) + M_{00}p_{01|0}\right)\cdot\left(M_{01}(1 - p_{01|1}) + M_{11}p_{10|1}\right)}$$

where we skipped indexes $i, j$ of counter variables and error probabilities for brevity, i.e., $M_u := M_u(j), M_{uv} := M_{uv}(ij), p_{uv|w} := p_{uv|w}(ij)$. Then the optimal decision to activate a component of the retrieval output $\hat{u}$ is

$$\hat{u}_j = 1 \quad \text{if} \quad x_j := b_j + \sum_{i=1}^{n} w_{ij}\tilde{u}_i \geq 0$$

where the "dendritic potential" $x_j$ corresponds to the log-odds-ratio $\log\left(\mathrm{pr}\left[u_j^{\mu} = 1 \middle| \tilde{u}\right] / \mathrm{pr}\left[u_j^{\mu} = 0 | \tilde{u}\right]\right)$. For further specifics of auto-association and an elaborate disucussion of the relation between BOM and BCPNN see [Knoblauch, 2024]. For the experiments of this study we used fixed noise estimates $p_{uv|w}(ij) := p_{uv}$ ignoring the conditioning on w. For example, for k active units per memory pattern with 10 percent add and miss noise, we simply used constant $p_{01} = 0.1k/(n-k)$ and $p_{10} = 0.1$ for all $i, j$.

Note that, unless $p_{01} = p_{10}$, the resulting weights are asymmetric, $w_{ij} \neq w_{ji}$, contrasting with the other learning rules from Table 2. However, for memory-efficient implementations, one may compute the weights "on-the-fly" during retrieval from the (symmetric) counter variables. As indicated above, this requires stroring only $\simeq \frac{n^2}{2}$ integers, similar as for the other learning rules.

Appendix B – Stochastic bisection method for estimating the number of stored patterns giving 90% correct recall

The pseudo-code used for efficient estimate of the crossing with the 90% level is given below. P0 was estimated from the theoretical max capacity with random binary patterns. The estimated value and spread is mean and standard deviation of the last P of each run.

```
dir = 0 ; dirs = [] ; P = P0 ; d = 10% P0
while (len(dirs)<20 or abs(mean(dirs))>0.1) :
        corr% = simulate(P,90) # Run the networks with P patterns
        dir_old = dir
        dir = sign(corr% - 90)
        P += dir * d
        if d>1 and dir*dir_old<0 :
           d = int(max(1,k*d + 0.5))
        elif d==1 :
           dirs := list of last 20 dir
 return P
```

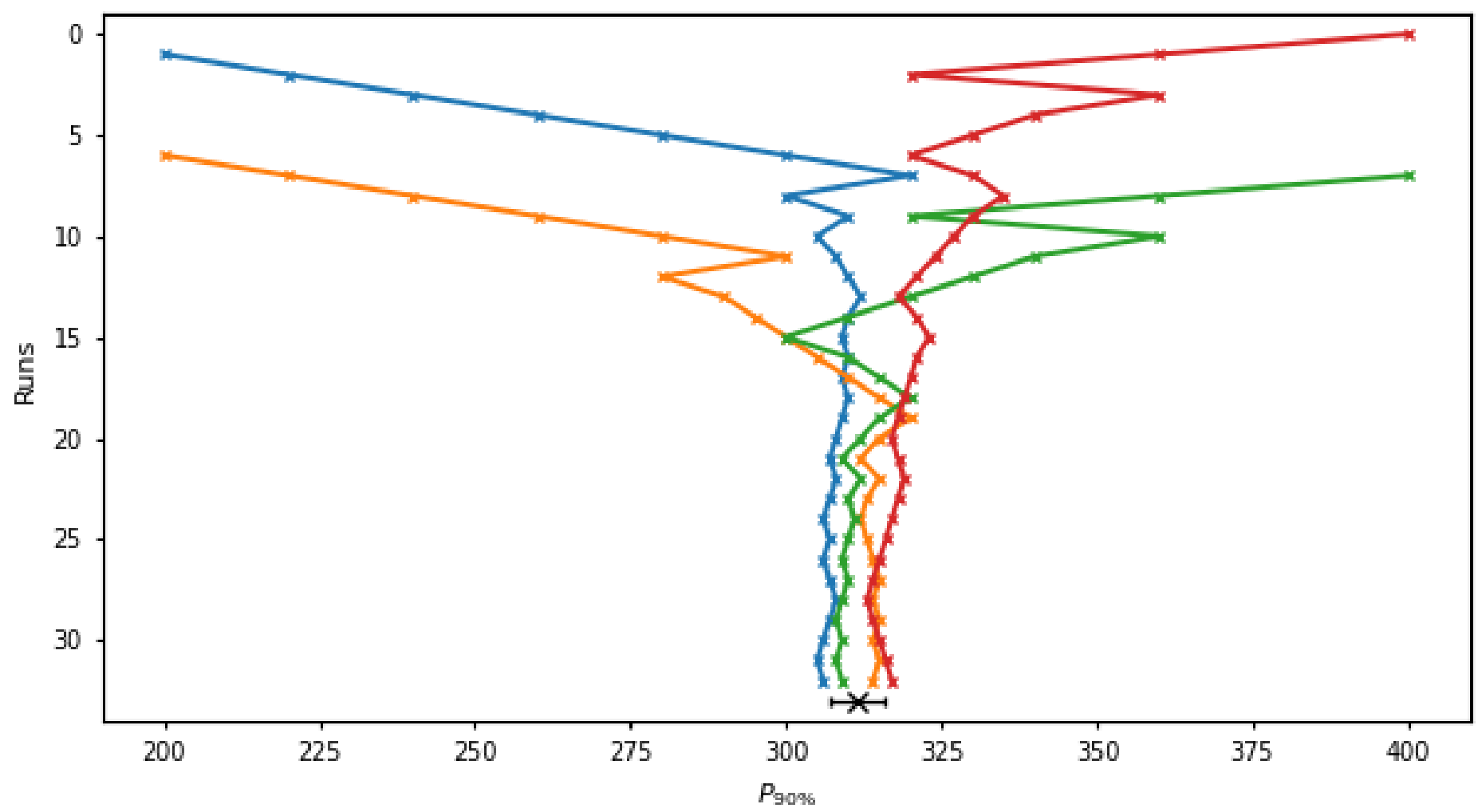

**Figure S1: Estimating storage capacity.** Output from the bisection method is shown with number of runs vs estimated $P_{90\%}$ for four runs of a BCP 16x16 network. P0 was 200 or 400 and the pattern distortion in each run resulted from resampling 2/16 hypercolumns. The resulting estimate mean was 312 and standard deviation 4.3.

## Appendix C – Data tables

| | | | | | | | | | | | | | |
|---|---|---|---|---|---|---|---|---|---|---|---|---|---|
| | | **KofN** | **binrand** | | | | | **HxM** | **binrand** | | | | |
| | N | 64 | 121 | 196 | 324 | 576 | 1024 | 64 | 121 | 196 | 324 | 576 | 1024 |
| WILL | mean | 12 | 30 | 61 | 142 | 330 | 796 | 24 | 64 | 132 | 266 | 586 | 1 275 |
| WILL | std | 0,94 | 1,12 | 2,17 | 2,92 | 1,89 | 3,74 | 1,12 | 2,32 | 1,74 | 2,59 | 4,58 | 2,87 |
| HEBB | mean | 10 | 14 | 40 | 74 | 156 | 381 | 12 | 30 | 43 | 106 | 233 | 578 |
| HEBB | std | 1 | 1,25 | 1,41 | 2,2 | 2,92 | 2,81 | 0,94 | 1,7 | 1,41 | 4 | 4,53 | 5,53 |
| HOPF | mean | 15 | 33 | 65 | 145 | 324 | 762 | 19 | 42 | 92 | 192 | 459 | 1 088 |
| HOPF | std | 1,63 | 0,83 | 1,74 | 2,62 | 1,83 | 7,59 | 1,25 | 2,04 | 1,72 | 2,38 | 2,32 | 3,28 |
| COV | mean | 12 | 42 | 65 | 144 | 326 | 752 | 17 | 37 | 95 | 195 | 466 | 1 088 |
| COV | std | 0,94 | 2,04 | 3,3 | 2,41 | 1,77 | 5,45 | 2,59 | 1,41 | 3,41 | 2,42 | 2,27 | 4,79 |
| PRCOV | mean | 14 | 41 | 74 | 154 | 350 | 791 | 19 | 46 | 105 | 219 | 488 | 1 109 |
| PRCOV | std | 1,48 | 2,69 | 1,12 | 2,94 | 3,1 | 4,46 | 1,25 | 2,13 | 2,17 | 2,83 | 6,02 | 5,07 |
| BCP | mean | 24 | 75 | 149 | 285 | 644 | 1 388 | 30 | 86 | 183 | 366 | 819 | 1 790 |
| BCP | std | 0,82 | 1,7 | 1,74 | 1,49 | 3,9 | 3,19 | 1,41 | 1,12 | 2,13 | 1,72 | 3,67 | 4,49 |
| BOM | mean | 39 | 93 | 160 | 296 | 602 | 1 318 | 33 | 86 | 169 | 342 | 756 | 1 634 |
| BOM | std | 1,33 | 2,06 | 3,51 | 1,76 | 7,72 | 3,64 | 0,83 | 1,12 | 3,9 | 3,44 | 5,45 | 6,88 |
| | | **KofN** | **silent** | | | | | **HxM** | **silent** | | | | |
| | N | 64 | 121 | 196 | 324 | 576 | 1024 | 64 | 121 | 196 | 324 | 576 | 1024 |
| WILL | mean | 20 | 33 | 54 | 88 | 129 | 210 | 26 | 38 | 61 | 106 | 161 | 247 |
| WILL | std | 1,12 | 1,58 | 2,68 | 1,8 | 2,59 | 2,53 | 1,41 | 2,69 | 2,17 | 2,83 | 4,98 | 2,76 |
| HEBB | mean | 10 | 12 | 14 | 17 | 23 | 17 | 10 | 14 | 16 | 19 | 23 | 18 |
| HEBB | std | 1 | 1 | 1 | 0,83 | 1,25 | 0,83 | 1 | 1,25 | 1,63 | 1,25 | 1,25 | 1,12 |
| HOPF | mean | 14 | 24 | 33 | 37 | 42 | 30 | 17 | 26 | 36 | 42 | 43 | 34 |
| HOPF | std | 0,83 | 1,25 | 1,7 | 1,41 | 1,41 | 1,41 | 0,83 | 1,41 | 2,04 | 1,98 | 2,69 | 1,41 |
| COV | mean | 10 | 19 | 28 | 42 | 58 | 80 | 12 | 24 | 36 | 54 | 71 | 102 |
| COV | std | 1 | 1,25 | 1,12 | 2,04 | 2 | 3,1 | 0,94 | 1,25 | 1,74 | 1,25 | 3,18 | 2,45 |
| PRCOV | mean | 19 | 30 | 44 | 92 | 137 | 239 | 19 | 41 | 67 | 110 | 204 | 335 |
| PRCOV | std | 1,25 | 1,7 | 2,04 | 2,87 | 2,64 | 5,54 | 1,25 | 1,41 | 1,17 | 1,41 | 4,35 | 3,2 |
| BCP | mean | 24 | 53 | 88 | 149 | 222 | 350 | 30 | 86 | 175 | 366 | 851 | 1 875 |
| BCP | std | 0,94 | 2,89 | 1,89 | 4,81 | 2,68 | 3,44 | 1,7 | 3,41 | 2,04 | 1,72 | 6,71 | 4,19 |
| BOM | mean | 31 | 68 | 120 | 201 | 302 | 461 | 30 | 86 | 166 | 331 | 736 | 1 593 |
| BOM | std | 1,25 | 1,12 | 2,42 | 2,36 | 3,25 | 5,15 | 1,41 | 1,12 | 5,59 | 3,19 | 3,53 | 7,4 |
| | | **KofN** | **cbinrand** | | | | | **HxM** | **cbinrand** | | | | |
| | N | 64 | 121 | 196 | 324 | 576 | 1024 | 64 | 121 | 196 | 324 | 576 | 1024 |
| WILL | mean | 8 | 17 | 37 | 68 | 129 | 258 | 17 | 38 | 75 | 136 | 235 | 492 |
| WILL | std | 1,25 | 0,83 | 2,69 | 1,48 | 2,59 | 1,57 | 0,83 | 2,69 | 1,17 | 2,64 | 2,54 | 3,92 |
| HEBB | mean | 0 | 6 | 17 | 19 | 41 | 74 | 6 | 14 | 20 | 30 | 48 | 86 |
| HEBB | std | 0 | 0,82 | 0,82 | 1,25 | 1,99 | 1,48 | 0,5 | 1,25 | 1,25 | 1,41 | 2,22 | 3,41 |
| HOPF | mean | 0 | 17 | 29 | 54 | 82 | 126 | 14 | 24 | 36 | 70 | 99 | 188 |
| HOPF | std | 0 | 1,87 | 1,48 | 1,48 | 3,06 | 2,27 | 1,48 | 1,12 | 3,1 | 3,91 | 1,49 | 3,52 |
| COV | mean | 0 | 12 | 36 | 57 | 125 | 226 | 6 | 24 | 38 | 83 | 179 | 395 |
| COV | std | 0 | 1 | 1,74 | 1,41 | 4,76 | 2,76 | 0,5 | 1,25 | 1,33 | 1,83 | 1,77 | 2,42 |
| PRCOV | mean | 12 | 12 | 55 | 91 | 196 | 372 | 15 | 37 | 72 | 149 | 275 | 634 |
| PRCOV | std | 1,8 | 1 | 1,25 | 1,77 | 6,87 | 2,69 | 1,63 | 1,41 | 1,17 | 4,81 | 1,49 | 2,93 |
| BCP | mean | 15 | 50 | 93 | 165 | 353 | 682 | 19 | 68 | 130 | 247 | 535 | 1 140 |
| BCP | std | 1,63 | 2,14 | 1,17 | 2,04 | 1,77 | 3,43 | 3,03 | 1,25 | 2,37 | 2 | 4,61 | 2,71 |
| BOM | mean | 32 | 68 | 112 | 187 | 357 | 694 | 26 | 64 | 121 | 252 | 514 | 1 088 |
| BOM | std | 1,12 | 1,25 | 1,7 | 2,04 | 4,13 | 7,53 | 1,41 | 2,32 | 1,49 | 2,76 | 9,53 | 6,04 |

**Fig 5 data table**

| | | KofN | binrand | | | | | HxM | binrand | | | | |
|---|---|---|---|---|---|---|---|---|---|---|---|---|---|
| | N | 64 | 121 | 196 | 324 | 576 | 1024 | 64 | 121 | 196 | 324 | 576 | 1024 |
| WILL | mean | 0 | 0 | 0 | 4 | 0 | 3 | 0 | 0 | 0 | 0 | 0 | 0 |
| WILL | std | 0 | 0 | 0 | 0,5 | 0 | 0 | 0 | 0 | 0 | 0 | 0 | 0 |
| HEBB | mean | 8 | 14 | 33 | 54 | 111 | 294 | 10 | 20 | 43 | 92 | 185 | 438 |
| HEBB | std | 0,5 | 1,25 | 1,7 | 1,25 | 2,42 | 2,2 | 1 | 1,12 | 1,41 | 2,87 | 2,04 | 4,44 |
| HOPF | mean | 17 | 36 | 59 | 121 | 272 | 680 | 16 | 37 | 85 | 186 | 419 | 958 |
| HOPF | std | 0,83 | 1,99 | 1,48 | 2,69 | 2,91 | 5,27 | 1 | 1,41 | 2,98 | 2,45 | 2,83 | 8,27 |
| COV | mean | 16 | 24 | 58 | 114 | 275 | 680 | 15 | 37 | 85 | 186 | 419 | 958 |
| COV | std | 1 | 1,12 | 4,15 | 2,42 | 1,49 | 5,27 | 1,63 | 1,41 | 2,98 | 1,41 | 4 | 8,27 |
| PRCOV | mean | 16 | 40 | 66 | 137 | 299 | 702 | 16 | 45 | 96 | 197 | 444 | 993 |
| PRCOV | std | 1 | 0,89 | 1,48 | 2,62 | 1,76 | 6,26 | 1 | 0,94 | 1,89 | 2,68 | 2,42 | 4,5 |
| BCP | mean | 30 | 74 | 132 | 246 | 528 | 1 137 | 36 | 90 | 174 | 332 | 718 | 1 553 |
| BCP | std | 1,48 | 1,89 | 1,74 | 2,14 | 1,98 | 5,39 | 0,94 | 1,12 | 1,41 | 2,81 | 3,45 | 6,85 |
| BOM | mean | 26 | 73 | 136 | 257 | 538 | 1 145 | 24 | 58 | 119 | 247 | 568 | 1 275 |
| BOM | std | 1,41 | 1,17 | 3,18 | 3,19 | 3,02 | 3,26 | 0,82 | 2 | 1,41 | 2 | 2,98 | 2,71 |
| | | KofN | silent | | | | | HxM | silent | | | | |
| | N | 64 | 121 | 196 | 324 | 576 | 1024 | 64 | 121 | 196 | 324 | 576 | 1024 |
| WILL | mean | 0 | 0 | 3 | 0 | 0 | 0 | 0 | 0 | 6 | 10 | 0 | 0 |
| WILL | std | 0 | 0 | 0 | 0 | 0 | 0 | 0 | 0 | 0,5 | 1 | 0 | 0 |
| HEBB | mean | 10 | 14 | 13 | 14 | 18 | 14 | 10 | 14 | 14 | 17 | 23 | 16 |
| HEBB | std | 1 | 1,63 | 1,63 | 1,48 | 1,17 | 1 | 1 | 1,25 | 1 | 0,83 | 1,25 | 1,63 |
| HOPF | mean | 14 | 24 | 27 | 33 | 37 | 26 | 17 | 26 | 40 | 42 | 42 | 34 |
| HOPF | std | 1,48 | 1,25 | 1,41 | 1,58 | 1,41 | 1,48 | 0,83 | 1,41 | 1,41 | 1,74 | 1,98 | 1,41 |
| COV | mean | 10 | 19 | 27 | 33 | 53 | 70 | 12 | 24 | 36 | 52 | 66 | 93 |
| COV | std | 1 | 1,25 | 1,41 | 1,58 | 2,17 | 1,41 | 0,94 | 1,25 | 1,74 | 2,14 | 1,67 | 1,72 |
| PRCOV | mean | 16 | 26 | 43 | 84 | 131 | 214 | 17 | 41 | 61 | 102 | 172 | 311 |
| PRCOV | std | 1 | 1,12 | 1,41 | 1,89 | 1,74 | 8,07 | 0,83 | 1,41 | 2,17 | 2,17 | 3,14 | 3,2 |
| BCP | mean | 30 | 37 | 54 | 54 | 35 | 26 | 34 | 70 | 130 | 247 | 496 | 1 076 |
| BCP | std | 1,7 | 2,05 | 2 | 1,48 | 2,04 | 1,48 | 1,41 | 1,17 | 2,37 | 3,08 | 4,26 | 6,23 |
| BOM | mean | 20 | 6 | 92 | 157 | 260 | 406 | 24 | 58 | 108 | 216 | 464 | 976 |
| BOM | std | 1,12 | 0,82 | 1,8 | 2,69 | 3,28 | 3,2 | 1,84 | 2,68 | 1,41 | 4,35 | 2,49 | 4,66 |
| | | KofN | cbinrand | | | | | HxM | cbinrand | | | | |
| | N | 64 | 121 | 196 | 324 | 576 | 1024 | 64 | 121 | 196 | 324 | 576 | 1024 |
| WILL | mean | 0 | 0 | 0 | 0 | 0 | 0 | 0 | 0 | 0 | 0 | 0 | 0 |
| WILL | std | 0 | 0 | 0 | 0 | 0 | 0 | 0 | 0 | 0 | 0 | 0 | 0 |
| HEBB | mean | 0 | 6 | 16 | 14 | 30 | 54 | 4 | 13 | 17 | 30 | 44 | 82 |
| HEBB | std | 0 | 0,82 | 1,63 | 1,48 | 1,41 | 2 | 0,5 | 2,24 | 0,83 | 1,48 | 1,41 | 1,89 |
| HOPF | mean | 4 | 14 | 26 | 36 | 70 | 119 | 14 | 24 | 36 | 68 | 92 | 186 |
| HOPF | std | 0,5 | 1,25 | 1,12 | 0,94 | 1,49 | 2,83 | 1,48 | 1,12 | 1,74 | 1,48 | 1,89 | 2,13 |
| COV | mean | 0 | 9 | 28 | 53 | 119 | 198 | 0 | 22 | 40 | 84 | 164 | 367 |
| COV | std | 0 | 1 | 1,12 | 2,24 | 2,58 | 4,22 | 0 | 1,17 | 2,05 | 3,32 | 2,85 | 2,87 |
| PRCOV | mean | 0 | 14 | 48 | 77 | 168 | 339 | 13 | 32 | 64 | 142 | 257 | 588 |
| PRCOV | std | 0 | 1,25 | 0,83 | 1,49 | 3,7 | 2,91 | 1,48 | 1,6 | 2,32 | 3,02 | 4,5 | 6,38 |
| BCP | mean | 20 | 37 | 72 | 82 | 84 | 67 | 22 | 58 | 99 | 180 | 372 | 820 |
| BCP | std | 1,12 | 1,74 | 1,17 | 1,8 | 1,72 | 1,17 | 1,12 | 2,17 | 2,42 | 4,31 | 2,91 | 4,03 |
| BOM | mean | 24 | 4 | 20 | 4 | 4 | 3 | 17 | 45 | 88 | 181 | 383 | 836 |
| BOM | std | 1,25 | 0,5 | 7,42 | 0,5 | 0,5 | 0 | 0,83 | 0,94 | 2,81 | 2,04 | 8,55 | 7 |

**Fig 9 data table**